\documentclass[11pt,a4paper]{article}
\usepackage{emnlp2022}
\usepackage{times}
\usepackage{multirow}
\usepackage{multicol}
\usepackage{latexsym}

\usepackage{booktabs} 
\usepackage[normalem]{ulem}
\usepackage{xcolor} 
\usepackage{algorithm}
\usepackage[noend]{algpseudocode}
\usepackage{amsmath}
\usepackage{mathrsfs}
 \usepackage{amssymb}
\usepackage{subfigure}
\usepackage{makecell}
\usepackage{mathtools}
\usepackage[font=rm]{caption}
\DeclareCaptionType{copyrightbox}
\usepackage{shortvrb}
\usepackage{tabularx}
\usepackage{verbatim}
\usepackage{xspace}
\usepackage{listings}
\usepackage{tikz}
\usepackage{verbatim}
\usetikzlibrary{trees}
\usepackage{forest}
\forestset{
    nice empty nodes/.style={
        for tree={
            s sep=0.1em, 
            l sep=0.33em,
            inner ysep=0.4em, 
            inner xsep=0.05em,
            l=0,
            calign=midpoint,
            fit=tight,
            where n children=0{
               tier=word,
               minimum height=1.25em,
            }{},
            where n children=2{
               l-=1em,
            }{},
            parent anchor=south,
            child anchor=north,
            delay={if content={}{
                    inner sep=0pt,
                    edge path={\noexpand\path [\forestoption{edge}] 
                    			(!u.parent anchor) 
                               -- (.south)\forestoption{edge label};}
                }{}}
        },
    },
}
\usepackage{fontawesome}
\usepackage[multiple]{footmisc}
\usepackage[all]{nowidow}
\usepackage{balance}
\usepackage{microtype}

\usepackage{xspace}
\newcommand{\bert}{\textsc{Bert}\xspace}
\newcommand{\gumbel}{\textsc{Gumbel}\xspace}

\usepackage{gnuplottex}
\usepackage[medium,compact]{titlesec}
\usepackage{enumitem}

\DeclareMathOperator*{\pop}{pop}
\DeclareMathOperator*{\len}{len}
\DeclareMathOperator*{\push}{push}
\DeclareMathOperator*{\KL}{KL}

\setlist{itemsep=0pt,parsep=0pt}

\newcommand{\Patk}[1]{\mbox{\Pat@$#1$}}
\newcommand{\RBPatp}[1]{\mbox{\RBP@$#1$}}

\newcommand{\NDCGatk}[1]{\mbox{\NDCG@$#1$}}
\newcommand{\ERRatk}[1]{\mbox{\ERR@$#1$}}

\newcommand{\myurl}[1]{{\url{#1}}}

\newcommand{\mycomment}[1]{}

\newlength{\onedigit}
\newcounter{todocount}
\title{Fast-R2D2:\vspace{-0.2ex} \\
A Pretrained Recursive Neural Network based on Pruned CKY\vspace{-0.2ex} \\
for Grammar Induction and Text Representation\vspace{-0.85ex}}

\date{}

\author{Xiang Hu\footnotemark[2] \quad Haitao Mi\footnotemark[2]~\thanks{~Work done while at Ant Group. To contact Haitao, haitaomi@global.tencent.com} \quad Liang Li\footnotemark[3] \quad Gerard de Melo\footnotemark[4] \\
 Ant Group\footnotemark[2] \\
 \tt \{aaron.hx, haitao.mi\}@alibaba-inc.com\footnotemark[2] \\
 School of Cyber Science and Technology, Shandong University, China /\\
 Key Laboratory of Cryptologic Technology and \\
 Information Security of Ministry of Education, Shandong University /\\Quancheng Laboratory, China\footnotemark[3]\\ 
 \tt li.liang@sdu.edu.cn\footnotemark[3] \\
 Hasso Plattner Institute / University of Potsdam\footnotemark[4] \\
 \tt gdm@demelo.org\footnotemark[4]  \\}

\begin{document}
\maketitle
\begin{abstract}
Chart-based models have shown great potential in unsupervised grammar induction, running recursively and hierarchically, 
but requiring $O(n^3)$ time-complexity.
The Recursive Transformer based on Differentiable Trees (R2D2) makes it possible to scale to large language model pretraining
even with a complex tree encoder, by introducing a heuristic pruning method.
However, its rule-based pruning process suffers from local optima and slow inference. 
In this paper, we propose a unified R2D2 method that overcomes these issues. 
We use a top-down unsupervised parser as a model-guided pruning method, 
which also enables parallel encoding during inference.
Our parser casts parsing as a split point scoring task by first scoring all split points for a given sentence and then 
using the highest-scoring one to recursively split a span into two parts.
The reverse order of the splits is considered as the order 
of pruning in the encoder.
We optimize the unsupervised parser by minimizing the Kullback–Leibler distance between tree probabilities from the parser and the R2D2 model.
Our experiments show that 
our Fast-R2D2 significantly improves the grammar induction 
quality and achieves competitive results in downstream 
tasks.\footnote{The code is available at: \url{https://github.com/alipay/StructuredLM\_RTDT}}
\end{abstract}

\section{Introduction}
Compositional, hierarchical, and recursive processing are widely believed to be essential traits of human language across diverse  linguistic theories~\cite{DBLP:journals/tit/Chomsky56,chomsky2014aspects}.
Chart-based models~\cite{DBLP:journals/corr/MaillardCY17,kim-etal-2019-compound,dblp:conf/naacl/drozdovvyim19,hu-etal-2021-r2d2} have made promising progress in both grammar induction and hierarchical encoding in recent years.
The differential CKY encoding architecture of \newcite{DBLP:journals/corr/MaillardCY17} simulates 
the hierarchical and recursive process explicitly by introducing an energy function to combine all possible derivations when constructing each cell representation.
However, this entails a cubic time complexity, which makes it impossible to scale to large 
language model training like BERT~\cite{devlin2018}. 
Simultaneously, its cubic memory cost also limits the tree encoder's ability to draw on huge parameter models as a backbone.

\newcite{hu-etal-2021-r2d2} introduced a heuristic pruning method, successfully reducing the time complexity 
to a linear number of compositions.
The experiments show that chart-based models exhibit great potential for grammar induction and representation learning when applying a sophisticated tree encoder such as Transformers with large corpus pretraining, leading to a Recursive Transformer based on Differentiable Trees, or R2D2 for short.
However, R2D2's heuristic pruning approach is rule-based and only considers certain composition probabilities.
Thus, trees constructed in this way are not guaranteed to be globally optimal. Moreover, as each step during pruning is based on previous decisions, the entire encoding process is sequential and thus slow in the inference stage.

In this work, we resolve these issues by proposing a unified method with a new global pruning strategy based on a lightweight and fast top-down parser. 
We cast parsing as split point scoring, where we first encode the input sentence with a bi-directional LSTM, and score all split points in parallel.
Specifically, for a given sentence, the parser first scores each split point between words in parallel by looking at its left and right contexts, and then recursively splits a span (starting with the whole sentence) into two sub-spans by picking the highest-scoring split point among the current split candidates.
Subsequently, the reverse order of the sorted split points can serve as the merge order to guide the pruning of the CKY encoder, which enables the encoder to search for more reasonable trees.
As the gradient of the pretrained component cannot be back-propagated to the parser, inspired by URNNG~\cite{dblp:conf/naacl/kimrykdm19}, we optimize the parser by taking trees sampled from the CKY chart table generated by the encoder as ground truth. Thus, the parser and the chart-based encoder promote each other in this way back and force just like the strategy and value networks in AlphaZero~\cite{DBLP:journals/nature/SilverSSAHGHBLB17}.
Additionally, the pretrained tree encoder can compose sequences recursively in parallel according to the trees generated by the parser, which makes Fast-R2D2 a Recursive Neural Network~\cite{DBLP:journals/ai/Pollack90,DBLP:conf/emnlp/SocherPWCMNP13} variant.




In this paper, we make the following main contributions:
\begin{enumerate}
    \item We propose an architecture to jointly pretrain parser and encoder of a recursive network in linear memory cost. Experiments show that our pretrained parser outperforms models custom-tailored for grammar induction.
    \item By encoding in parallel following trees generated by the top-down parser, Fast-R2D2 significantly improves the inference speed 30 to 50 fold compared to R2D2.
    \item We pre-train Fast-R2D2 on a large corpus and evaluate it on downstream tasks. The experiments demonstrate that a pretrained recursive model based on an unsupervised parser significantly outperforms pretrained sequential Transformers~\cite{DBLP:conf/nips/VaswaniSPUJGKP17} with the same parameter size in single sentence classification tasks.
\end{enumerate}

\section{Preliminaries}

\subsection{R2D2 Architecture}\label{sec:r2d2}

\begin{figure}[htb!]
  \centering
  \includegraphics[width=0.45\textwidth]{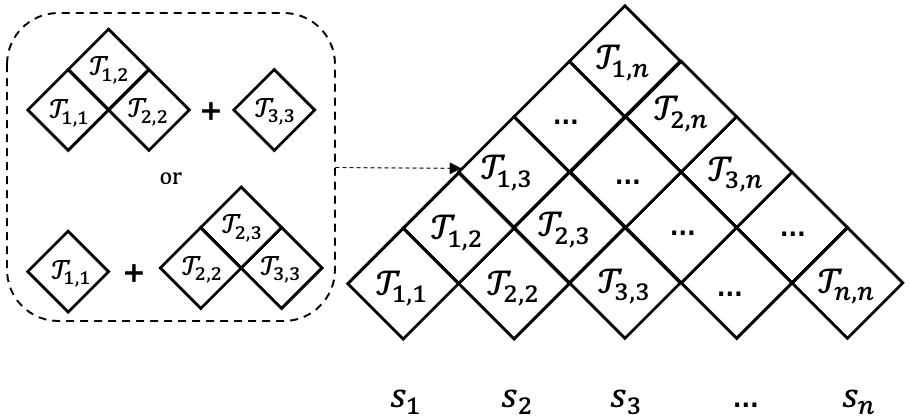}
  \caption{Chart data structure. There are two alternative ways of generating $\mathcal{T}_{1,3}$:
     combining either ($\mathcal{T}_{1,2}$, $\mathcal{T}_{3,3}$) or ($\mathcal{T}_{1,1}$, $\mathcal{T}_{2,3}$).}
  \label{fig:chart_data}
\end{figure}

\paragraph{Differentiable Trees.} 
R2D2 follows the work of \newcite{DBLP:journals/corr/MaillardCY17} in defining a CKY-style~\cite{10.5555/1097042,kasami1966efficient,younger1967recognition} encoder.
For a sentence $\mathbf{S} = \{s_{1}, s_{2},..., s_{n}\}$ with $n$ words or word-pieces, 
it defines a chart table as illustrated in Figure~\ref{fig:chart_data}. In the table, each cell $\mathcal{T}_{i, j}$ is a tuple $\langle e_{i, j}, p_{i, j}, \widetilde{p}_{i,j} \rangle$, where
$e_{i, j}$ is a vector representation, $p_{i, j}$ is the probability of a single composition step, 
and $\widetilde{p}_{i,j}$ is the probability of the subtree for the span $[i, j]$ over the sub-string $s_{i:j}$.
When $i$ equals $j$, the table has terminal nodes $\mathcal{T}_{i, i}$ with $e_{i, i}$ initialized with the embeddings of input tokens $s_{i}$, while
$p_{i, i}$ and $\widetilde{p}_{i,i}$ are set to one. 
When $j>i$, the representation $e_{i, j}$ is a weighted sum of intermediate combinations $c_{i, j}^{k}$, defined as: 
\begin{align}
\label{eq:tree_encoder}
&c_{i,j}^{k}, \  p_{i,j}^{k} = f(e_{i, k}, e_{k+1, j})\\
\label{eq:tree_prob}
&\widetilde{p}_{i,j}^{k} = p_{i,j}^{k} \,\, \widetilde{p}_{i,k} \,\, \widetilde{p}_{k+1,j}\\
&\boldsymbol{\alpha}_{i,j} = \text{\gumbel} (\log( \mathbf{\widetilde{p}}_{i,j}))\\
&e_{i,j} =  [c_{i,j}^{i}, c_{i,j}^{i+1}, ..., c_{i,j}^{j-1}]\,\boldsymbol{\alpha}_{i,j}\\
&[p_{i,j},\widetilde{p}_{i,j}] = \boldsymbol{\alpha}_{i,j}^\intercal [\boldsymbol{p}_{i,j}, \boldsymbol{\widetilde{p}}_{i,j}]
\end{align}
$k$ is a split point from $i$ to $j-1$, $f(\cdot)$ is an $n$-layer Transformer encoder,
$p_{i,j}^{k}$ and $\widetilde{p}_{i,j}^{k}$ denote the single step combination probability and the subtree probability, respectively, at split point $k$,
$\boldsymbol{p}_{i,j}$ and $\boldsymbol{\widetilde{p}}_{i,j}$ are the concatenation of all $p_{i,j}^{k}$ or $\widetilde{p}_{i,j}^{k}$ 
values, 
and \gumbel is the Straight-Through Gumbel-Softmax operation of \newcite{DBLP:conf/iclr/JangGP17} with temperature set to one. As \gumbel picks the optimal splitting point $k$ at each cell in practice,
it is straightforward to recover the complete derivation tree from the root node $\mathcal{T}_{1,n}$ in a top-down manner recursively.

\begin{figure*}[htb!]
    \flushleft
    \includegraphics[width=1\textwidth]{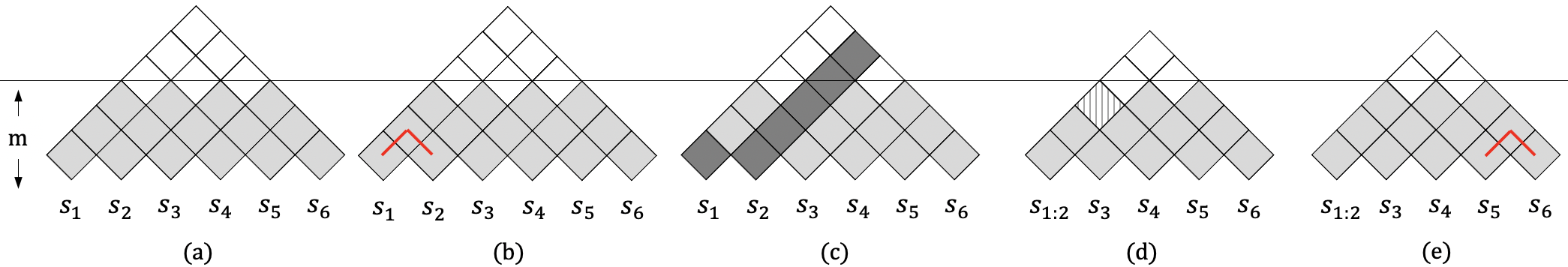}
    \vspace{-3ex}
    \caption{Example of chart pruning and encoding process. With R2D2's original heuristic pruning, cells to merge are selected according to local composition probabilities. For better model-based pruning, we propose selecting cells according to the merge order estimated by a top-down parser.}\vspace{-1ex}
    \label{fig:pruning}
\end{figure*}

\paragraph{Heuristic pruning.} 
As shown in Figure~\ref{fig:pruning}, R2D2 starts to prune if all cells beneath height $m$ have been encoded. 
The heuristic rules work as follows:
\begin{enumerate}
    \item Recover the maximum sub-tree for each cell at the $m$-th level, and collect all cells at the $2$nd level that appear in any sub-tree.
    \item Rank candidates in Step 1 by the composition probability $p_{i, j}$, and pick the highest-scoring cell as a non-splittable span (e.g., $\mathcal{T}_{1,2}$).
    \item Remove any invalid cells that would break the now non-splittable span from Step 2, e.g., the dark cells in (c), and reorganize the chart table much like in the Tetris game as in (d).
    \item Encode the blank cells at the $m$-th level, e.g., the cell highlighted with stripes in (d), and go back to Step 1 until the root cell has been encoded.
\end{enumerate}


\paragraph{Pretraining.}


To learn meaningful structures without gold trees, \newcite{hu-etal-2021-r2d2} propose a self-supervised pretraining objective. Similar to the bidirectional masked language model task, R2D2 reconstructs a given token $s_i$ based on its context representation $e_{1,i-1}$ and $e_{i+1, n}$. The probability of each token is estimated by the tree encoder defined in R2D2. The final objective is:
\begin{equation}
\label{eq:bilm_loss}
\underset{\theta}{\mathrm{min}}\,\sum_{i=1}^{n} -\log\,p_{\theta}(s_{i} \mid e_{1:i-1}, e_{i+1:n})
\end{equation}
\section{Methodology}

\subsection{Global Pruning Strategy}
We propose a top-down parser based on syntactic distance~\cite{DBLP:conf/acl/BengioSCJLS18} to evaluate scores for all split points in a sentence and generate a merge order according to the scores. 

\begin{figure}[htb!]
  \centering
  \includegraphics[width=0.4\textwidth]{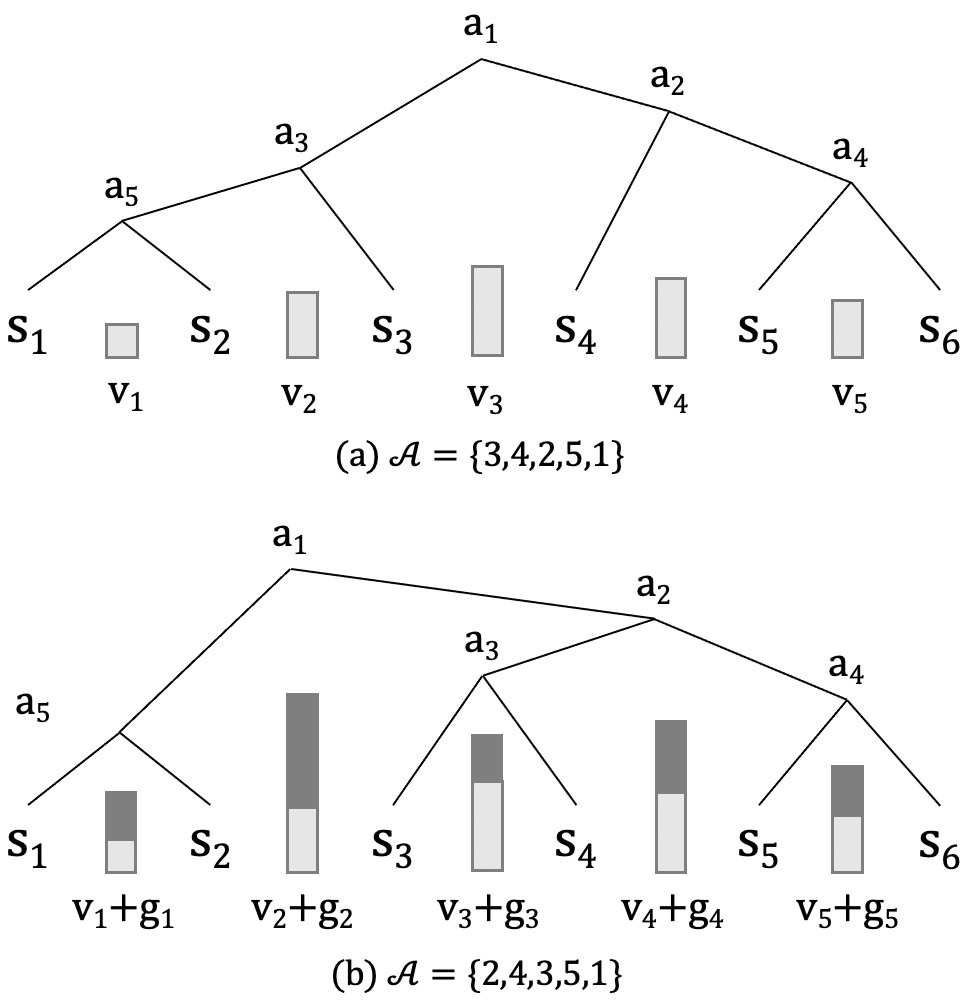}
  \caption{(a) A parsed tree obtained by sorting split scores ($v_i$). (b) A sampled tree by adding Gumbel noise ($g_i$ in dark vertical bars).}
  \label{fig:sample_demo}
\end{figure}

\paragraph{Top-down parser.}
Given a sentence $\mathbf{S} = \{s_{1}, s_{2},..., s_{n}\}$, there are $n-1$ split points between words. 
We define a top-down parser by giving confidence scores to all split points as follows:
\begin{equation}
\begin{aligned}
\textbf{v}=[v_{1}, v_{2}, ..., v_{n-1}] = f(\mathbf{S};\theta)
\end{aligned}
\end{equation}
To keep it simple and rigorously maintain linear complexity, we select bidirectional LSTMs as the backbone, though Transformers are also an option.
As shown in Figure~\ref{fig:sample_demo}, 
first, a bi-directional LSTM encodes the sentence, 
and then, for each split point, an MLP over the concatenation of the left and right context representations yields the final split scores. 
Formally, we have:
\begin{equation}
\begin{aligned}
\label{eq:split_score_eval}
&\overrightarrow{\textbf{h}},\overleftarrow{\textbf{h}} = \mathrm{BiLSTM}(\mathbf{E};\theta )\\
&v_{i} = \mathrm{LayerNorm}(\mathrm{MLP}(\overrightarrow{\textbf{h}}_{i}\oplus \overleftarrow{\textbf{h}}_{i+1}))
\end{aligned}
\end{equation}
Here, $\mathbf{E}$ is the embedding of the input sentence $\mathbf{S}$, while
$\overrightarrow{\textbf{h}}$ and $\overleftarrow{\textbf{h}}$ denote the forward and reverse representation, respectively. 
$v_{i}$ is the score of the $i$-th split point, whose left and right context representations are $\overrightarrow{\textbf{h}}_{i}$ and $\overleftarrow{\textbf{h}}_{i+1}$. 
Given scores $[v_1, v_2, ..., v_{n-1}]$, one can easily recover the binary tree shown in Figure~\ref{fig:sample_demo}:
We recursively split a span (starting with the entire sentence) into two sub-spans by picking the split point with the highest score in the current span.
Taking the sentence in Figure~\ref{fig:sample_demo} (a) as an 
example, we split the overall sentence at split point $3$ in the first step, which leads to two sub-trees over $s_{1:3}$ and $s_{4:6}$. 
Then we split $s_{1:3}$ at $2$ and $s_{4:6}$ at $4$. We can continue this procedure until the complete tree has been recovered. 

\paragraph{Tree sampling.}
In the training stage, we perform sampling over the computed scores $[v_1, v_2, ..., v_{n-1}]$ in order to increase the robustness and exploration of our model.
Let $\mathcal{P}^{t}$ denote the list of split points at time $t$ in ascending order, which is $\{1,2,3,...,n\!-\!1\}$ in the first step. 
Then a particular split point $a_{t}$ is selected from $\mathcal{P}^{t}$ by sampling based on the probabilities estimated by stacking of split points scores. The sampled
$\{a_{1}, a_{2}, ..., a_{n-1}\}$ together form the final split point sequence $\mathcal{A}$. 
At each time step, we remove $a_{t}$ from $\mathcal{P}^{t}$ when $a_{t}$ is selected, then sample the next split point until the set of remaining split points is empty. Formally, we have:
\begin{align}
&a_{t} \sim \mathrm{softmax}(\textbf{v}^t)\\
&\mathcal{P}^{t+1}=\mathcal{P}^{t}\setminus \{{a_{t}}\}
\end{align}
where $\textbf{v}^t$ is concatenation of $v_{i}$ in $\mathcal{P}^{t}$.
As the Gumbel-Max trick \cite{GUMBEL,DBLP:conf/nips/MaddisonTM14} provides a simple and efficient way
to draw samples from a categorical distribution with class probabilities, we can obtain $a_{t}$ via the Gumbel-Max trick as:
\begin{align}
&a_{t} = \underset{i}{\mathrm{argmax}}\,[v_{i}+g_{i}], i \in \mathcal{P}^{t},
\end{align}
where $g_{i}$ is the Gumbel noise for the $i$-{th} split point. 
Therefore, the aforementioned process is equivalent to sorting the original sequence of split points scores with added Gumbel noise.
Figure~\ref{fig:sample_demo} (b) shows a sampled tree with respect to the split point scores.
The split point sequence $\mathcal{A}$ can hence be obtained simply as:
\begin{align}
\mathcal{A} = \underset{i}{\mathrm{argsort}}(\textbf{v} + \textbf{g})
\end{align}
Here, $\mathrm{argsort}$ sorts the array in descending order and returns the indices of the original array.
The sampled $\mathcal{A}$ is $\{2, 4, 3, 5, 1\}$ in Figure~\ref{fig:sample_demo} (b).

\paragraph{Span Constraints.}
As word-pieces~\cite{wu2016google} and Byte-Pair Encoding (BPE) are commonly used in pretrained language models,
it is straightforward to incorporate multiple word-piece constraints into the top-down parser to reduce word-level parsing errors.
We denote a list of span constraints composed of beginning and end positions of non-split-table spans as $\mathcal{C}$, defined as $\mathcal{C}=\{(b_{1}, e_{1}), (b_{2}, e_{2}), ..., (b_{n}, e_{n})\}$. 
For each $(b_{i}, e_{i})$ in $\mathcal{C}$, there should be a sub-tree for a span covering the sub-string $s_{b_{i}:e_{i}}$. 
This goal can be achieved by simply adjusting the scores of all splits within the spans in C by $-\delta$. To make them smaller than the scores of span boundaries, $\delta$ could be defined as $(\max(\textbf{v}) - \min(\textbf{v}) + c)$, where $c$ could be any positive number.

\paragraph{Model-based Pruning.}
We denote the reverse order of the split point sequence $\mathcal{A}$ as $\mathcal{M}$ and
then treat $\mathcal{M}$ as a bottom-up merge order inferred by the top-down parser based on the global context.
Subsequently, the complete pruning process is as follows: 
\begin{enumerate}
\item Pick the next merge index by invoking Alg~\ref{alg:next_merge_point}. 
\item Perform Steps 3 and 4 in the heuristic pruning part in Section~\ref{sec:r2d2}
\end{enumerate}
As shown in Figure~\ref{fig:pruning}, we still retain the threshold and the pruning logic of R2D2, 
but we select cells to merge according to $\mathcal{M}$ instead of following heuristic rules. 
Specifically, given a shrinking chart table, 
we select the next merge index among the second row by popping and modifying $\mathcal{M}$ in Algorithm~\ref{alg:next_merge_point}.

\begin{algorithm}[!h]
\small
    \caption{Next merge index in the second row}
    \label{alg:next_merge_point}
    \begin{algorithmic}[1] 
        \Function{Next-Index}{$\mathcal{M}$}
        \State{$i = \pop(\mathcal{M})$}\Comment{Index}
        \For{$j \in 1$ to $\mathcal{M}.\mathrm{len}$}
        \If {$\mathcal{M}_{j} > i$} \Comment{Merging at left}
        \State {$\mathcal{M}_{j} = \mathcal{M}_{j} - 1$} \Comment{Shift left}
        \EndIf
        \EndFor
        \State{\Return{$i$}}
        \EndFunction
    \end{algorithmic}
\end{algorithm}

Take the example in Figure~\ref{fig:sample_demo} (b) for instance: 
$\mathcal{M}$ starts with $\{1, 5, 3, 4, 2\}$.
Then we merge the first cell in the second row in Figure~\ref{fig:pruning} (b),
and obtain a new $\mathcal{M} = \{4, 2, 3, 1\}$. 
In the next round, we treat the 4th cell covering $s_{5:6}$ as a non-splittable cell in Figure~\ref{fig:pruning} (e), 
and $\mathcal{M}$ becomes $\{2, 3, 1\}$.

\subsection{Optimization}\label{sec:opt}

We denote the tree probabilities estimated by the top-down parser and R2D2 as $p_{\theta}(\textbf{z}|\textbf{S})$, $q_{\phi}(\textbf{z}|\textbf{S})$, respectively. 
The difficulty here is that while
$q_{\phi}(\textbf{z}|\textbf{S})$ may be optimized by the objective defined in Equation~\ref{eq:bilm_loss}, 
there is no gradient feedback for $p_{\theta}(\textbf{z}|\textbf{S})$. 
To make $p_{\theta}(\textbf{z}|\textbf{S})$ learnable, an intuitive solution is to fit $p_{\theta}(\textbf{z}|\textbf{S})$ to $q_{\phi}(\textbf{z}|\textbf{S})$ by minimizing their Kullback–Leibler distance. 
While the tree probabilities of both distributions are discrete and not exhaustive,
inspired by URNNG~\cite{dblp:conf/naacl/kimrykdm19}, a Monte Carlo estimate for the gradient with respect to $\theta$ can be defined as:
\begin{equation}
\small
\begin{aligned}
&\triangledown_{\theta} \KL[q_{\phi}(\textbf{z}|\textbf{S}) \parallel p_{\theta}(\textbf{z}|\textbf{S}) ] \\
= &\triangledown_{\theta} \mathbf{E}_{z \sim q_{\phi}(\textbf{z}|\textbf{S})}[\log \frac{q_{\phi}(\textbf{z}|\textbf{S})}{p_{\theta}(\textbf{z}|\textbf{S})}] \\
\approx &-\triangledown_{\theta} \frac{1}{K}\sum_{k=1}^{K}\log p_{\theta}(\textbf{z}^{(k)}|\textbf{S})
\end{aligned}
\vspace{-1.5pt}
\end{equation}
with samples $\textbf{z}^{(1)}$, ..., $\textbf{z}^{(K)}$ from $q_{\phi}(\textbf{z}|\textbf{S})$. 
Algorithm~\ref{alg:sample} shows the complete sampling process from $q_{\phi}(\textbf{z}|\textbf{S})$.
Specifically, we sample split points with corresponding span boundaries recursively as in previous work \cite{DBLP:journals/corr/cmp-lg-9805007,DBLP:conf/emnlp/FinkelMN06,dblp:conf/naacl/kimrykdm19} 
with respect to the intermediate tree probabilities calculated during hierarchical encoding.

\begin{algorithm}[!h]
\small
    \caption{Top-down tree sampling for R2D2}
    \label{alg:sample}
    \begin{algorithmic}[1] 
        \Function{Sample}{$\mathcal{T}_{1,n}$} \Comment{Root cell}
        \State {$Q = [\mathcal{T}_{1,n}]$}
        \State {$K = []$}
        \While {$Q$ is not empty}
        \State{$\mathcal{T} = \pop(Q)$}
        \State {$i,j = \mathcal{T}.i$, $\mathcal{T}.j$} \Comment{Start/end indices}
        \State{$L = \mathcal{T}.\mathrm{splits}$} \Comment{$m$ splits at most}
        \State{$\tau=0$}
        \For {$k \in 1$ to $\len(L)$}
        \State {$w_{k} = \widetilde{p}_{i,j}^{L[k]}$} \Comment{Using Equation~\ref{eq:tree_prob}}
        \State {$\tau = \tau + w_{k}$} \Comment {Sum up all $w_{k}$}
        \EndFor
        \State{$idx \sim \mathrm{Cat}([w_{1}/\tau, ..., w_{\len(L)}/\tau])$} \\
        \Comment{Sample a split point}
        \State{$\push(K, (L[idx], i, j))$} \\
        \Comment{Keep the split point and span boundary}
        \If {$L[idx] > i$} \Comment{Add left child}
            \State{$\push(Q, \mathcal{T}_{i, L[idx]})$}
        \EndIf
        \If {$L[idx] + 1 < j$} \Comment{Add right child}
            \State{$\push(Q, \mathcal{T}_{L[idx]+1, j})$}
        \EndIf
        \EndWhile
        \State{\Return{$K$}}
        \EndFunction
    \end{algorithmic}
\end{algorithm}

A sequence of split points and corresponding spans is returned by the sampler. For the $k$-{th} sample $\textbf{z}^{(k)}$, let $p_{\theta}(a_{t}^{k}|\textbf{S})$ denote the probability of taking $a_{t}^{k}$ as split from span $(i_{t}^{k}, j_{t}^{k})$ at the $t$-{th} step. Formally, we have:
\begin{equation}
\small
\begin{aligned}
p_{\theta}(a_{t}^{k}|\textbf{S}) &= \frac{e^{v_{a^{k}_{t}}}}{e^{v_{i^{k}_{t}}} + ... + e^{v_{j^{k}_{t}}}} \\
\log p_{\theta}(\textbf{z}^{(k)}|\textbf{S}) &= \sum_{t=1}^{n-1} \log p_{\theta}(a_{t}^{k}|\textbf{S}),
\end{aligned}
\end{equation}
where $i_{t}^{k}$ and $j_{t}^{k}$ denote the start and end of the corresponding span. Please note here that the $v_i$ are not adjusted by span constraints. 

\subsection{Downstream Tasks}
\label{sec:downstream}
\paragraph{Inference.}
In this paper, we mainly focus on classification tasks as downstream tasks. We consider the root representation as representing the entire sentence.
As we have two models pre-trained in our framework -- an R2D2 encoder and a top-down parser -- we have two ways of generating the representations:
\begin{enumerate}
    \item[a)] Run forced encoding over the binary tree from the top-down parser with the R2D2 encoder.
    \item[b)] Use the binary tree to guide the pruning of the R2D2 encoder, and take the root representation $e_{1,n}$.
\end{enumerate}
It is obvious that the first approach is much faster than the latter one, as the R2D2 encoder only runs $n-1$ times in forced encoding, 
and can run in parallel layer by layer, e.g., we may run compositions at $a_5$, $a_3$, and $a_4$ in parallel in Figure~\ref{fig:sample_demo} (b).
We explore both of these approaches in our experiments.

\paragraph{Training Objectives.}
As suggested in prior work \cite{radford2018improving,howard-ruder-2018-universal,gururangan-etal-2020-dont}, 
given a pretrained model, continued pretraining on an in-domain corpus with the same pretraining objective can yield a better generalization ability.
Thus, we simply combine our pretraining objectives via summation in all downstream tasks. At the same time, as the downstream task may guide R2D2 to more reasonable tree structures, we still maintain the KL loss to enable the parser to continuously update.
For the two inference methods,
we uniformly select the root representation $e_{1,n}$ as the representation for a given sentence followed by an MLP, and estimate the cross-entropy loss, denoted as $\mathcal{L}_\mathrm{forced}$ and $\mathcal{L}_\mathrm{cky}$, respectively. Let $\mathcal{L}_\mathrm{KL}$ denote the KL loss described in Section~\ref{sec:opt} and $\mathcal{L}_\mathrm{bilm}$ denote the bidirectional language model loss described in Eq~\ref{eq:bilm_loss}.
The final loss is:
\begin{equation}
\mathcal{L} = \mathcal{L}_\mathrm{forced} + \mathcal{L}_\mathrm{cky} + \mathcal{L}_\mathrm{bilm} + \mathcal{L}_\mathrm{KL}
\end{equation}

\section{Experiments}
\label{sec:exps}
\subsection{Unsupervised Grammar Induction}

\subsubsection{Setup}\label{sec:LM_setup}
\paragraph{Baselines and Evaluation.} 
For comparison, we include six recent strong models for unsupervised parsing with available open source implementations: StructFormer \cite{DBLP:conf/acl/ShenTZBMC20}, Ordered Neurons~\cite{DBLP:conf/iclr/ShenTSC19}, URNNG~\cite{dblp:conf/naacl/kimrykdm19}, DIORA~\cite{dblp:conf/naacl/drozdovvyim19}, C-PCFG~\cite{kim-etal-2019-compound}, and R2D2~\cite{hu-etal-2021-r2d2}. 
To observe the marginal gain from pretraining, we also include Fast-R2D2 without pretraining denoted as Fast-R2D2$_{\rm w/o}$.
Following~\newcite{htut-etal-2018-grammar}, we train all systems on a training set consisting only of raw text, and evaluate and report the results on an annotated test set. 
As an evaluation metric, we adopt sentence-level unlabeled $F_1$ computed using the script from \newcite{kim-etal-2019-compound}.
We compare against the non-binarized gold trees per convention.
The results of Fast-R2D2 are obtained from 3 runs of each model with different random seeds in pre-training.
The best checkpoint for each system is picked based on scores on the validation set. 
Fast-R2D2 is pretrained with span constraints for the word level but without span constraints for the word-piece level.
To support word-piece level evaluation, 
we convert gold trees to word-piece level trees 
by simply breaking each terminal node into a non-terminal node with its word-pieces as terminals, e.g., (NN discrepancy) into (NN (WP disc) (WP \#\#re) (WP \#\#pan) (WP \#\#cy)).

\paragraph{Environment.} EFLOPS~\cite{DBLP:conf/hpca/DongCZYWFZLSPGJ20} is a highly scalable distributed training system designed by Alibaba. With its optimized hardware architecture and co-designed supporting software tools, including ACCL~\cite{DBLP:journals/micro/DongWFCPTLLRGGL21} and KSpeed (the high-speed data-loading service), it could easily be extended to 10K nodes (GPUs) with linear scalability.

\paragraph{Hyperparameters.} The tree encoder of our model uses 4-layer Transformers with 768-dimensional embeddings, 
3,072-dimensional hidden layer representations, and 12 attention heads. 
The top-down parser of our model uses a 4-layer bidirectional LSTM with 128-dimensional embeddings and 256-dimensional hidden layer. The sampling number $K$ is set to be 256.
Training is conducted using Adam optimization with weight decay using a learning rate of $5 \times 10^{-5}$ for the tree encoder and $1 \times 10^{-2}$ for the top-down parser.
The batch size is set to 64 per GPU for $m$=$4$, though we also limit the maximum total length for each batch, such that excess sentences are moved to the next batch. The limit is set to 1,536. It takes about 120 hours for 60 epochs of training with $m$=$4$ on 8 A100 GPUs.

\paragraph{Data.}  For English, to fully leverage the scalability of Fast-R2D2, we pretrain Fast-R2D2 on WikiText103~\cite{DBLP:conf/iclr/MerityX0S17}
and then fine-tune the model on the Penn Treebank (PTB)~\cite{marcus-etal-1993-building}
for 10 epochs with the same objective.
WikiText103 is split at the sentence level, and sentences longer than 200 after tokenization are discarded (about 0.04‰ of the original data). 
The total number of sentences is 4,089,500, and the average sentence length is 26.97.
For Chinese, we use a subset of Chinese Wikipedia (Simplified Characters) for pretraining, specifically the first 10,000,000 sentences shorter than 150 characters and then fine-tune on Chinese Penn Treebank (CTB) 8.0~\cite{ctb8}.
We test our approach on PTB WSJ data with the standard splits (2--21 for training, 22 for validation, 23 for test) and the same preprocessing as in recent work \cite{kim-etal-2019-compound}, where we discard punctuation and lower-case all tokens. 
To explore the universality of the model across languages, we further evaluate using the CTB,
on which we also remove punctuation.
Note that in all settings, the training and fine-tuning is conducted entirely on raw unannotated text.

\subsubsection{Results and Discussion}

\begin{table}
\newcommand{\invzero}{\hphantom{0}}
\begin{center}
\setlength{\tabcolsep}{3.pt}
\resizebox{0.45\textwidth}{!}{
\begin{tabular}{@{}l|l|l|l|l@{}}
                    &  eval & mem. & \multicolumn{1}{c|}{WSJ}  & \multicolumn{1}{c}{CTB}  \\
Model               & gran. & cplx  &  $F_1(\mu)$ & $F_1(\mu)$\\ \hline \hline
Left Branching (W)  & WD & $O(n)$& \invzero 8.15  & 11.28 \\
Right Branching (W) & WD & $O(n)$& 39.62 & 27.53 \\
Random Trees (W)    & WD & $O(n)$ & 17.76 & 20.17 \\
\hline
URNNG (W)           & WD & $O(n^3)$& 45.4$^\dag$ & ~~--- \\
ON-LSTM (W)         & WD & $O(n)$  & 47.7$^\dag$ & 24.73 \\
DIORA (W)           & WD & $O(n^3)$& 51.4 & ~~---  \\
StructFormer (W)    & WD & $O(n^2)$& 54.0$^\ddagger$ & ~~--- \\
C-PCFG (W)          & WD & $O(n^3)$& 55.2$^\dag$ & 49.95 \\ \hline
R2D2 (WP)           & WD & $O(n)$ & 48.11 & 44.85  \\
Fast-R2D2$^*$(W)$_{\rm w/o}$ & WD & $O(n)$ & 48.24 & 45.24 \\
Fast-R2D2$^*$(WP)$_{\rm w/o}$ & WD & $O(n)$ & 48.89 & 45.26 \\
Fast-R2D2$^*$(WP)  & WD & $O(n)$ & \textbf{57.22} & \textbf{53.13} \\
\hline \hline
R2D2 (WP)           & WP & $O(n)$  & 52.28 & 63.94 \\ 
Fast-R2D2(WP)      & WP & $O(n)$ & 50.20 & \textbf{67.79} \\
Fast-R2D2$^*$(WP)  & WP & $O(n)$& \textbf{53.88} & 67.74 \\ \hline
\end{tabular}
}
\end{center}
\caption{Unsupervised parsing results with words (W) or word-pieces (WP) as input. ``eval gran." is short for evaluation granularity.
        Values marked with $^{\dag}$ are taken from \newcite{kim-etal-2019-compound}, while $^{\ddagger}$ denotes values taken from \newcite{DBLP:conf/acl/ShenTZBMC20}.
        The bottom three systems are all pre-trained or trained 
        at the word-piece level \textbf{without} span constraints and are measured against word-piece level golden trees. ${\rm w/o}$ means without pretraining.}
\label{tbl:constituency_parsing}
\end{table}

Table~\ref{tbl:constituency_parsing} shows the results of all systems with words (W) and word-pieces (WP) as input on the WSJ and CTB test sets. 
When we evaluate all systems on word-level golden trees, 
our Fast-R2D2 performs substantially better than R2D2 across both datasets.
We denote as Fast-R2D2 the method of using the parser to guide the pruning and selecting the best tree using the chart table and as Fast-R2D2$^*$ the system that uses the top-down parser for tree induction with subsequent R2D2 encoding.
Interestingly, the results suggest that Fast-R2D2$^*$ outperforms Fast-R2D2, especially on the WSJ test set.
Additionally, pretrained Fast-R2D2$^*$
outperforms the models specifically designed for grammar induction.

\begin{table}[!htb]
\small
\begin{center}
\setlength{\tabcolsep}{3.5pt}
\resizebox{0.48\textwidth}{!}{ %
\begin{tabular}{@{}ll| l l l l l l@{}}
 & Model  & WD & NNP & VP & SBAR\\\hline \hline
\multirow{5}{*}{\rotatebox[origin=c]{90}{WSJ}} & DIORA (WP)  & 94.63 & 77.83 & 17.30 & 22.16\\
& C-PCFG (W)                  & ~~--- & ~~--- & 41.7$^\dag$ & 56.1$^\dag$ \\
& C-PCFG (WP)                  & 87.35 & 66.44 & 23.63 & 40.40 \\
& R2D2 (WP)    & \textbf{99.76} & \textbf{86.76} & 24.74 & 39.81\\
& Fast-R2D2$^*$ (WP) & 97.67 & 83.44 & \textbf{63.80} & \textbf{65.68} \\ \hline \hline
\multirow{3}{*}{\rotatebox[origin=c]{90}{CTB}} & C-PCFG(WP) &89.34 & 46.74 & 39.53 & ~~---\\
 & R2D2 (WP) & 97.16 & 67.19 & 37.90 & ~~---\\
 & Fast-R2D2$^*$ (WP) & \textbf{97.80} & \textbf{68.57} & \textbf{46.59} & ~~---
 \\ \hline \hline
\end{tabular}
}
\end{center}
\caption{Recall of constituents and words. WD means word.  Values with $^{\dag}$ are taken from \newcite{kim-etal-2019-compound}.}
\label{tbl:unsupervised_chunking}
\end{table}

Following \newcite{dblp:conf/naacl/kimrykdm19} and \newcite{drozdov-etal-2020-unsupervised},
we also compute the recall of constituents when evaluating on word-piece level golden trees.
Besides standard constituents, we also compare the recall of word-piece chunks and proper noun chunks. 
Proper noun chunks are extracted by finding adjacent unary nodes with the same parent and tag NNP. 
Table~\ref{tbl:unsupervised_chunking} reports the recall scores for constituents and words on the WSJ and CTB test sets. 
Compared with the R2D2 baseline, 
our Fast-R2D2 performs slightly worse for small semantic units, 
but significantly better over larger semantic units (such as VP and SBAR) on the WSJ test set.
On the CTB test set, our Fast-R2D2 outperforms R2D2 on all constituents. 

From Tables~\ref{tbl:constituency_parsing}~and~\ref{tbl:unsupervised_chunking}, 
we conclude that Fast-R2D2 overall obtains better results than R2D2 on CTB, while faring slightly worse than R2D2 only for small semantic units on WSJ. We conjecture that this difference stems from differences in  tokenization between Chinese and English. 
Chinese is a character-based language without complex morphology, where collocations of characters are consistent with the language, making it easier for the top-down parser to learn them well. 
In contrast, word-pieces for English are built based on statistics, and individual word-pieces are not necessarily natural semantic units. Thus, there may not be sufficient semantic self-consistency, such that it is harder for a top-down parser with a small number of parameters to fit it well.

\subsection{Downstream Tasks}
We next consider the effectiveness of Fast-R2D2 in downstream tasks. This experiment is not intended to advance the state-of-the-art on the GLUE benchmark but rather to assess to what extent our approach performs respectably against the dominant inductive bias as in conventional sequential Transformers.

\subsubsection{Setup}
\paragraph{Data and Baseline.}
We fine-tune pretrained models on several datasets,
including SST-2, CoLA, QQP, and MNLI from the GLUE benchmark~\cite{wang2018glue}.
As sequential Transformers with their dominant inductive bias remain the norm for numerous NLP tasks, 
we mainly compare Fast-R2D2 with \bert~\cite{devlin2018} as a representative pretrained model based on a sequential Transformer. 
We did not include recursive models such as Gumbel-Tree-LSTMs~\cite{DBLP:conf/aaai/ChoiYL18} and CRvNN~\cite{DBLP:conf/icml/ChowdhuryC21} among our baselines, as they are not pretrained models.
In order to compare the two forms of inductive bias fairly and efficiently,
we pretrain \bert models with 4 layers and 12 layers as well as our Fast-R2D2 from scratch on the WikiText103 corpus following Section~\ref{sec:LM_setup}. 
Considering that longer inputs in the pre-training stage are helpful for BERT’s downstream task performance, we use the original corpus that is not split into sentences as inputs.
For simplicity, Fast-R2D2 is fine-tuned without span constraints.
Following the common settings, we add an MLP layer over the root representation of the R2D2 encoder for single-sentence classification. 
For cross-sentence tasks such as QQP and MNLI, we feed the root representations of the two sentences into the pretrained tree encoder of R2D2 as left and right inputs, 
and also add a new task ID as another input term to the R2D2 encoder. 
Then we feed the hidden output of the new task ID into another MLP layer to predict the final label.
We train all systems across the four datasets for 10 epochs 
with a learning rate of $5\times 10^{-5}$, batch size $64$, and maximum input length $200$.
We validate each model in each epoch and report the best results on development sets.

\begin{table}
\begin{center}
\setlength{\tabcolsep}{1.5pt}
\resizebox{0.48\textwidth}{!}{
\begin{tabular}{l|c|r r|r r}
\multirow{4}{*}{Model} & \multirow{4}{*}{Para.} & \multicolumn{2}{c|}{Single sent.} & \multicolumn{2}{c}{Cross sent.} \\
 &  & \begin{tabular}[c]{@{}l@{}}SST-2\\ (Acc.)\end{tabular} & \begin{tabular}[c]{@{}l@{}}CoLA\\ (Mcc.)\end{tabular} & \begin{tabular}[c]{@{}l@{}}QQP\\ (F1)\end{tabular} & \begin{tabular}[c]{@{}l@{}}MNLI\\m/mm\\ (Acc.)\end{tabular}            \\ \hline \hline
\bert (4L)  & 52M & 84.98 & 17.07 & 84.01 & 73.73/74.63 \\
\bert (12L) & 116M & 90.25 & 40.72 & 87.13 & 80.00/80.41 \\ \hline
R2D2        & 52M & 89.33 & 34.79 & 84.27 &  69.35/68.72 \\ \hline
Fast-R2D2$^\dag$& {\multirow{2}{*}{\begin{tabular}[c]{@{}c@{}}\\52M/\\ 10M\end{tabular}}} & 87.50 & 8.67 & 83.97 & 69.53/69.50 \\
Fast-R2D2$^*\dag$& {} & 88.30 & 10.14 & 84.07 & 69.36/69.11 \\
Fast-R2D2  & {} & 90.25 & 38.45 & 84.35 & 69.36/68.80 \\ 
Fast-R2D2$^*$& {} & 90.71 & 40.11 & 84.32 & 69.64/69.57\\
\hline \hline
\end{tabular}
}
\end{center}
\caption{Downstream results. All systems are pretrained from scratch on WikiText103.
        Para.\ describes the number of parameters for each model. Fast-R2D2 contains the R2D2 encoder and top-down parser, two components with 52M and 10M parameters, respectively.
        Mcc.\ stands for Matthew's correlation coefficient.
        Fast-R2D2 with $\dag$ are models fine-tuned without $\mathcal{L}_\mathrm{bilm}$ for an ablation study.
    }\vspace{-10pt}
\label{tbl:classification}
\end{table}
\subsubsection{Results and Discussion}
Table~\ref{tbl:classification} shows the corresponding scores on SST-2, CoLA, QQPl, and MNLI. 
In terms of the parameter size, our Fast-R2D2 model has 52M and 10M parameters for the R2D2 encoder and top-down parser, respectively.
It is clear that 12-layer \bert is significantly better than 4-layer \bert.
As mentioned in Section~\ref{sec:downstream}, Fast-R2D2 has two options to construct the final tree and representation for a given input sentence:
Fast-R2D2$^*$ uses the output tree from the top-down parser, while Fast-R2D2 uses the best tree inferred by the R2D2 encoder.
Similar to the results for unsupervised parsing, Fast-R2D2$^*$ in classification tasks again outperforms Fast-R2D2.
We hypothesize that trees generated by the top-down parser without Gumbel noise are more stable and reasonable.
Fast-R2D2 significantly outperforms 4-layer \bert and achieves competitive results compared to 12-layer \bert in single sentence classification tasks such as SST-2 and CoLA, but still performs significantly worse in the cross-sentence tasks. 
We believe this is an expected result, as there is no cross-attention mechanism in the inductive bias of Fast-R2D2. 
However, the performance of Fast-R2D2 on classification tasks shows that the inductive bias of R2D2 has higher parameter utilization than sequentially applied Transformers.
Importantly, we demonstrate that a Recursive Neural Network variant with an unsupervised parser can achieve comparable results to pretrained sequential Transformers even with fewer parameters and interpretable intermediate results, 
Hence, our Fast-R2D2 framework provides an alternative for NLP tasks.

\subsection{Speed Evaluation}
To assess the time cost, we mainly compare sequential Transformers and Fast-R2D2 in forced encoding on various sequence length ranges. We randomly select 1,000 sentences for each range from WikiText103 and report the average time consumption on a single A100 GPU. \bert is based on the open source Transformers library\footnote{\url{https://github.com/huggingface/transformers}} and R2D2 is based on the official code in \newcite{hu-etal-2021-r2d2}.\footnote{\url{https://github.com/alipay/StructuredLM_RTDT/tree/r2d2}}

\begin{table}
\small
\begin{center}
\setlength{\tabcolsep}{3.pt}
\resizebox{0.45\textwidth}{!}{
\begin{tabular}{l|rrrr}
\multirow{2}{*}{Model} & \multicolumn{4}{c}{Sequence Length Ranges} \\\cline{2-5}
      & \multicolumn{1}{c|}{0--50} & \multicolumn{1}{l|}{50--100} & \multicolumn{1}{l|}{100--200} & 200--500 \\ 
\hline
\bert (12L) & \multicolumn{1}{r|}{1.36}     & \multicolumn{1}{r|}{1.46}       & \multicolumn{1}{r|}{1.62}        & 2.38 \\ \hline
R2D2  & \multicolumn{1}{r|}{38.06}     & \multicolumn{1}{r|}{173.74}       & \multicolumn{1}{r|}{555.95}        &    ---     \\
Fast-R2D2  & \multicolumn{1}{r|}{4.67} & \multicolumn{1}{r|}{14.91} & \multicolumn{1}{r|}{39.73} & 150.26 \\
Fast-R2D2* & \multicolumn{1}{r|}{1.28} & \multicolumn{1}{r|}{2.96}  & \multicolumn{1}{r|}{5.56}  & 10.70 \\ 
\hline \hline
\end{tabular}
}
\end{center}
\caption{Inference time in seconds for various systems to process 1,000 sentences with a batch size of 50.}
\label{tbl:speed_test}
\end{table}

Table~\ref{tbl:speed_test} shows the inference time in seconds for different systems to process 1,000 sentences with a batch size of 50.
Running R2D2 is time-consuming, since the heuristic pruning method involves substantial memory exchanges between GPU and CPU. 
In Fast-R2D2, we alleviate this problem by using model-guided pruning to accelerate the chart table processing,
in conjunction with a code implementation in CUDA, Fast-R2D2 reduces the inference time significantly. 
Fast-R2D2$^{*}$ further improves the inference speed by running forced encoding in parallel over the binary tree generated by the parser, which is about 30--50 times faster than R2D2 in various ranges. 
Although there is still a gap in speed compared to sequential Transformers, Fast-R2D2$^{*}$ is sufficiently fast for most NLP tasks while producing interpretable intermediate representations.

\section{Related Work}
\label{sec:related_works}

Many attempts have been done in grammar induction and hierarchical encoding. \citet{DBLP:conf/conll/Clark01} and \citet{DBLP:conf/acl/KleinM02}
provided some of the first successful statistical approaches to grammar induction. There have been multiple recent papers that focus on structure induction based on language modeling objectives~\cite{DBLP:conf/nips/ShenTHLSC19,DBLP:conf/iclr/ShenTSC19,DBLP:conf/acl/ShenTZBMC20,kim-etal-2019-compound}. \newcite{DBLP:journals/ai/Pollack90} proposed to use RvNN as a recursive architecture to encode text hierarchically, and \newcite{DBLP:conf/emnlp/SocherPWCMNP13} showed the effectiveness of RvNNs with gold trees for sentiment analysis. In this work, we focus on models that are capable of learning meaningful structures in an unsupervised way and encoding text over the induced tree hierarchically.

In the line of work on learning a sentence representation with structures, \newcite{DBLP:conf/iclr/YogatamaBDGL17} jointly train their shift-reduce parser and sentence embedding components without gold trees.
As their parser is not differentiable, they have to resort to reinforcement training, resulting in increased variance, which may easily collapse to trivial left or right branching trees. Gumbel-Tree-LSTMs~\cite{DBLP:conf/aaai/ChoiYL18} construct trees by recursively selecting two terminal nodes to merge and learning the composition probability via downstream tasks. CRvNN~\cite{DBLP:conf/icml/ChowdhuryC21} makes the entire process end-to-end differentiable and parallel by introducing a continuous relaxation.
URNNG~\cite{dblp:conf/naacl/kimrykdm19} propose the first architecture to jointly pretrain parser and encoder based on RNNG \cite{dyer-etal-2016-recurrent}. However, it has $O(n^3)$ complexity and remains unable to improve upon a right-branching baseline when punctuation is removed.
\newcite{DBLP:journals/corr/MaillardCY17} propose an alternative approach, based on a differentiable CKY encoding. The algorithm is differentiable due to a soft-gating approach, which approximates discrete candidate selection by a probabilistic mixture of the constituents available in a given cell of the chart. While their work relies on annotated downstream tasks to learn structures, \newcite{dblp:conf/naacl/drozdovvyim19} propose a novel auto-encoder-like pretraining objective based on the inside--outside algorithm~\cite{Baker1979TrainableGF,DBLP:conf/icgi/Casacuberta94}. 

\section{Conclusion}
In this paper, we have presented Fast-R2D2, which improves the performance and inference speed of R2D2 by introducing a fast top-down parser to guide the pruning of the R2D2 encoder.
Pretrained on the same corpus, Fast-R2D2 significantly outperforms sequential Transformers with a similar scale of parameters on classification tasks. 
Experimental results show that Fast-R2D2 is a promising and feasible way to learn hierarchical text representations, which is different from layer stacking models and can also generate interpretable intermediate representations.
As future work, we are investigating leveraging the intermediate representations in additional downstream tasks.

\section{Limitations}
Our approach has three major limitations. First, Fast-R2D2 has shortcomings with regard to cross-sentence tasks due to the lack of cross-attention between sentences. Second, Fast-R2D2 requires greater memory resources for pretraining compared to sequential Transformers.  At each invocation, the composition function takes four inputs and runs on $m$ candidates, which means the total number of calls to the MLP is $4mn$. Hence, the pretraining time of  Fast-R2D2 is about 3 to 4 times that of BERT with 12 layers. Finally, our model does not beat most of the baselines in grammar induction when trained on WSJ only. A side effect of the pruning strategy is that the chart-table actually is a sparse table, which means not all tokens are reconstructed based on complete context.  This issue can be alleviated by pre-training on a large corpus, which is what our method is designed for, and why we introduce the ability to parallelize the computation.

\section{Acknowledgement}
We would like to thank the Aliyun EFLOPS team for their substantial support in designing and providing a cutting-edge training platform to facilitate fast experimentation in this work. We would also like to thank the Zhixiaobao team for their support in applying our model to real applications.

\newpage
\bibliographystyle{acl_natbib} 
\bibliography{anthology}

\onecolumn
\section{Appendix}
\subsection{Tree Examples}

\begin{figure}[htb!]
  \centering
  \includegraphics[width=0.99\textwidth]{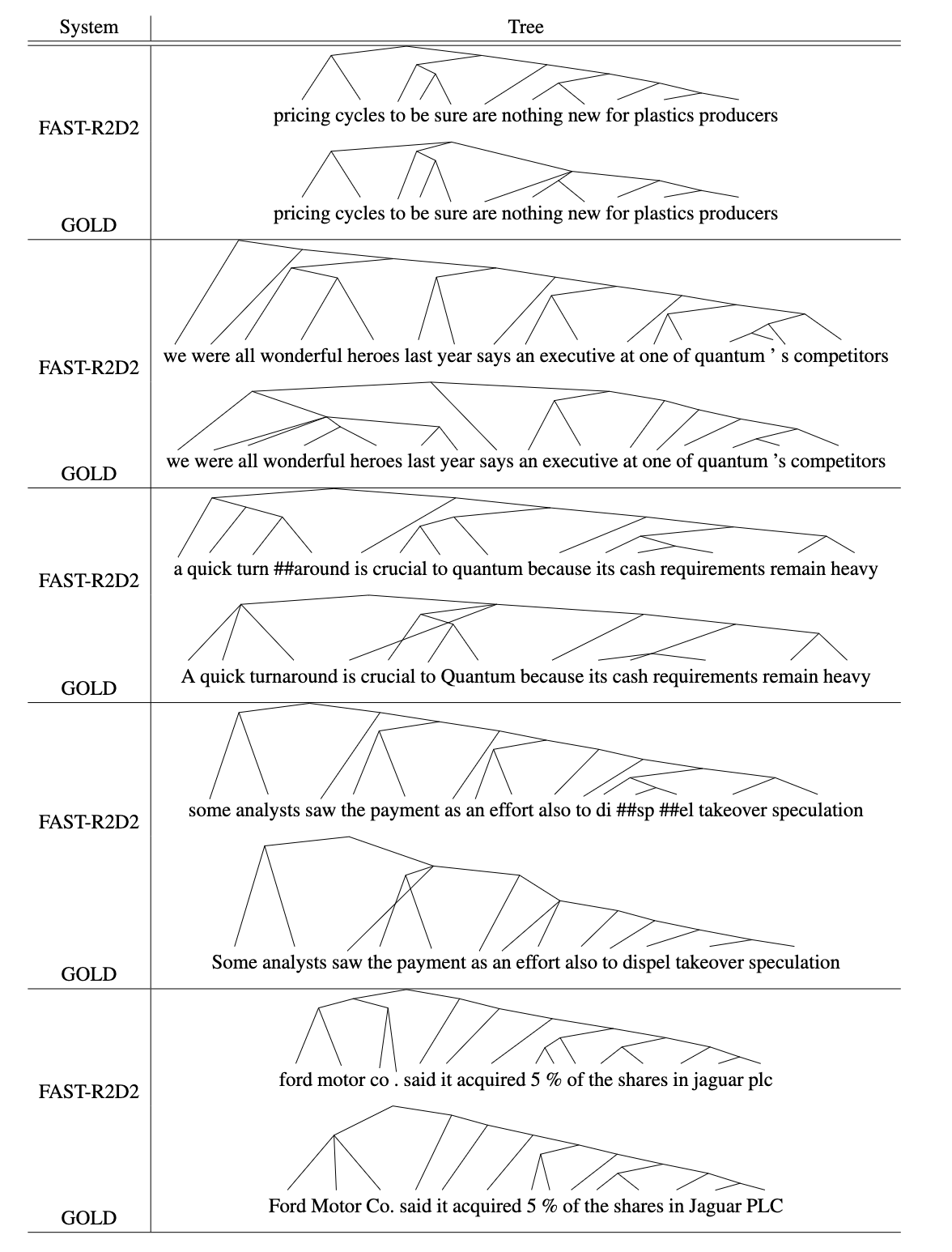}
\end{figure}

\end{document}